\documentclass[preprint,12pt,authoryear]{elsarticle}

\pdfoutput=1

\usepackage{hyperref}
\hypersetup{
  pdfauthor={Maciej Wielgosz and Stefano Puliti and Phil Wilkes and Rasmus Astrup}
}

\usepackage[acronym]{glossaries}
\usepackage{glossary-inline}
\setglossarystyle{inline}
\setacronymstyle{long-short}
\newacronym{DL}{DL}{Deep Learning}
\newacronym{LSTM}{LSTM}{Long Short-Term Memory}
\newacronym{GRU}{GRU}{Gated Recurrent Unit}
\newacronym{RMSE}{RMSE}{Root-Mean-Square Error}
\newacronym{RNN}{RNN}{Recurrent Neural Network}
\newacronym{ASIC}{ASIC}{Application-Specific Integrated Circuit}
\newacronym{FPGA}{FPGA}{Field-Programmable Gate Array}
\newacronym{FNN}{FNN}{Feed-forward Neural Network}
\newacronym{CNN}{CNN}{Convolutional Neural Network}

\newacronym{TLS}{TLS}{terrestrial laser scanning}
\newacronym{PLS}{PLS}{personalized laser scanning}
\newacronym{MLS}{MLS}{mobile laser scanning}
 
\makenoidxglossaries

\usepackage[range-phrase={--},range-units=single]{siunitx}
\mathchardef\mhyphen="2D 

\usepackage{booktabs}
\usepackage{tabularx}

\usepackage{easy-todo}

\usepackage{algorithm}
\usepackage{algpseudocode}
\usepackage{siunitx}

\journal{International Journal of Applied Earth Observation and Geoinformation}

\begin{document}

\begin{frontmatter}

\title{Point2Tree(P2T) - framework for parameter tuning of semantic and instance segmentation used with mobile laser scanning data in coniferous forest}

\author[1]{Maciej Wielgosz\corref{cor1}}
\ead{maciej.wielgosz@nibio.no}

\author[1]{Stefano Puliti}
\ead{stefano.puliti@nibio.no}

\author[2,3]{Phil Wilkes}
\ead{}

\author[1]{Rasmus Astrup}
\ead{rasmus.astrup@nibio.no}

\cortext[cor1]{Corresponding author}

\affiliation[1]{
    organization={Norwegian Institute for Bioeconomy Research (NIBIO), Division of Forest and Forest Resources},
    addressline={Høgskoleveien 8},
    postcode={1433},
    city={Ås},
    country={Norway}
}
\affiliation[2]{
    organization={Department of Geography, University College London},
    city={London},
    country={United Kingdom}
}
\affiliation[3]{
    organization={NERC National Centre for Earth Observation},
    city={Leicester},
    country={United Kingdom}
}

\begin{abstract}

This article introduces Point2Tree, a novel framework that incorporates a three-stage process involving semantic segmentation, instance segmentation, optimization analysis of hyperparemeters importance. It introduces a comprehensive and modular approach to processing laser points clouds in Forestry. We tested it on two independent datasets. The first area  was located in an actively managed boreal coniferous dominated forest in Våler, Norway, 16 circular plots of 400 square meters were selected to cover a range of forest conditions in terms of species composition and stand density. We trained a model based on Pointnet++ architecture which achieves 0.92 F1-score in semantic segmentation. As a second step in our pipeline we used graph-based approach for instance segmentation which reached F1-score approx. 0.6. The optimization allowed to further boost the performance of the pipeline by approx. 4 \% points.  
 
\end{abstract}


\begin{highlights}
\item A new modular framework for tuning  semantic and instance segmentation methods is demonstrated on point clouds obtained in boreal forests,
\item A new semantic segmentation model was developed specifically for boreal forests.
\item The new framework was demonstrated on data captured with a mobile laser scanning system.
\item We showed that it is possible to improve the segmentation quality through the optimization process, provided a sufficient and well structured dataset is provided,
\item Source code available at \href{https://gitlab.nibio.no/maciekwielgosz/instance_segmentation_classic}{https://gitlab.nibio.no/maciekwielgosz/instance\_segmentation\_classic}
\end{highlights}


\nonumnote{\printnoidxglossary[type=\acronymtype]}

\end{frontmatter}

\section{Introduction}
\label{section:intro}

The use of high-resolution 3D point clouds from \gls{TLS}, \gls{PLS}, and drone or helicopter-based laser scanning data have long been an area of intensive research for the characterization of forest ecosystems. In addition to measuring traditional variables such as stem volume, diameter, and height \cite{astrup2014approaches} or tree species \citep[e.g.][]{allen2022tree}, these very high-detail 3D forest structural data can allow new insight into single tree properties such as biomass \citep[e.g.][]{demol2022estimating, calders2015nondestructive}, stem curves \citep[e.g.][]{hyyppa2020accurate}, tree height growth \citep[e.g.][]{puliti2022tree}, wood quality \citep[e.g.][]{pyorala2018quantitative}, key ecological indicators \citep{calders2020terrestrial}, and phenotyping \citep{grubinger2020modeling,hartley2022assessing}. 

The emergence of improved sensor technology and implementation of Simultaneous Localization And Mapping (SLAM) algorithms have greatly improved the availability and reduced the cost of dense point clouds from mobile laser scanning (MLS) platforms.  The continuous move from stationary \gls{TLS} to mobile, personalized, or drone-based scanning systems has greatly increased the ease of scanning larger forest plots \citep{tockner2022automatic} and reduced the challenge of when not limited to fixed stations \citep[see][]{boucher2021sampling}. Recent studies have also pointed out that personalized or mobile laser devices may provide an improved \citep{donager2021adjudicating} or more cost-efficient \citep{kukenbrink2022benchmarking} alternative to traditional field measurements with calipers and hypsometers for collection of ground truth for air- or space-borne remote sensing. 

At the core of most forest applications of high-density 3D point clouds is the ability to efficiently, with high precision and accuracy, segment the point cloud into different compartments such as stem, leaf, or branches (semantic segmentation) and further into single tree point clouds, hereafter referred to as instance segmentation. Even though the development of segmentation algorithms has been a substantial field of research for many years, both semantic and instance segmentation remains a significant bottleneck to unleashing the full potential of high-density point clouds in the forest context. Most studies have relied on algorithmic approaches (e.g., clustering and circle fitting or voxel-based approaches \cite{wang2008lidar}) to identify and segment single trees \cite{vicari2019leaf,burt2019extracting}. In all cases, the segmentation routines tend to produce artifacts, and the single tree point clouds often require manual editing. In addition, such approaches are generally tailored to the specific data set and sensors they were developed on and are seldom transferable to new data.

Recent advances in the field of deep learning are triggering a new wave of studies looking into the possibilities to disentangle the complexity of high-density 3D point clouds and solve semantic \citep{krisanski2021sensor,hyyppa2020accurate}, instance segmentation \citep{windrim2020detection} and regression tasks \citep{oehmcke2021deep}. One promising avenue in this field is the development of sensor-agnostic models that can learn general point cloud features and allow their transferability independently from the characteristics of the input point cloud. The advantage of such models is that they can be used off-the-shelf on new data without hyperparameter tuning. One exemplary case of moving in the direction of sensor-agnostic models for forest point clouds is the study by \citet{krisanski2021sensor}, who developed a semantic segmentation model to classify primary features (i.e., ground, wood, and leaf) in forest 3D scenes captured with TLS, ALS, and MLS. While desirable, there are currently no sensor-agnostic tree instance segmentation models for point cloud data. However, steps have been made to integrate sensor-agnostic deep learning semantic models with more traditional algorithmic pipelines to solve the instance segmentation challenge \citep{krisanski2021sensor, oehmcke2022deep, chen2021individual}.

\textit{TLS2Trees} \citet{wilkes2022TLS2trees} addressed the instance segmentation problem while leveraging on the FSCT semantic segmentation model published by \citet{krisanski2021sensor}. In their approach, the wood classified points from the semantic segmentation are used to construct a graph through the point cloud, then uses a shortest path analysis to attribute points to individual stem bases\citet{wilkes2022TLS2trees}. In a final step, the leaf classified points are then added to each graph. A key aspect in this pipeline that affects the quality of the downstream products is the initial definition of the clusters done using the DBSCAN clustering method \citep{ester1996density}. The performance of DBSCAN depends on the separability of the instances, which is tightly linked to the output of the FSCT segmentation model (i.e., wood class parts, including stems and branches) and to the forest type. In particular, instance clustering can be challenging in dense forests with a substantial amount of woody branches in the lower parts of the crown (e.g. Norway spruce forests). One potential avenue to boost the quality of the initial definition of the instances is to develop new point cloud semantic segmentation models that allow for a clearer separability of the instances by, for example, focusing on the main tree stem (i.e., excluding branches). 

In \textit{TLS2Trees} \citet{wilkes2022TLS2trees} , the instance segmentation performance depends on a set of hyperparameters which should be individually tuned for a given type of forest to archive the best possible performance. So far, both in \textit{TLS2Trees} and other tree segmentation approaches tuning of this types of hyperparameters have traditionally been done manually by individual researchers for each data set through trial and error processes. However, the possibility for a systematic and automated approach for hyperparemeter optimization exists. Several potential methods that could solve the challenge exists (e.g. simulated annealing or Bayesian optimization \cite{brochu2010tutorial}). Furthermore, it is also worth keeping in mind that the gradient-based methods may be ill-posed due to the non-convex profile of the hyperparameter space. 


Leveraging on resent advances in semantic segmentation \cite{krisanski2021sensor} and instance segmentation approaches \cite{wilkes2022TLS2trees} this study introduces Point2Tree which is a new modular framework for semantic and instance segmentation for MLS data. The Point2Tree has two main modules (1) a newly trained Pointnet++ based semantic segmentation model with the classes optimized for coniferous forest (p2t\_semantic, see Tab.\ref{tab:pipeline-structure}), and (2) an optimisation procedure for instance segmentation hyperparameter optimization based on the Bayesian flow approach \cite{burt2019extracting}. Point2Tree is modular in the way that each of the components can be easily replaced by an improved module. We evaluate the performance of  both the semantic and instance segmentation of Point2Tree with settings including: (a) with and without hyperparameter optimization and (b) with both the new semantic Pointnet++ model (i.e. p2t\_semantic)  as well as with the semantic model from FSCT \cite{krisanski2021sensor} i.e. fsct\_semantic (see Tab.\ref{tab:pipeline-structure}). The evaluation is done against a newly annotated dataset from our study area as well as an existing independent dataset from another part of Europe.    

\section{Materials}

\subsection{Study area}
\begin{figure}[htbp]
    \centering
    \includegraphics[width=0.8\textwidth]{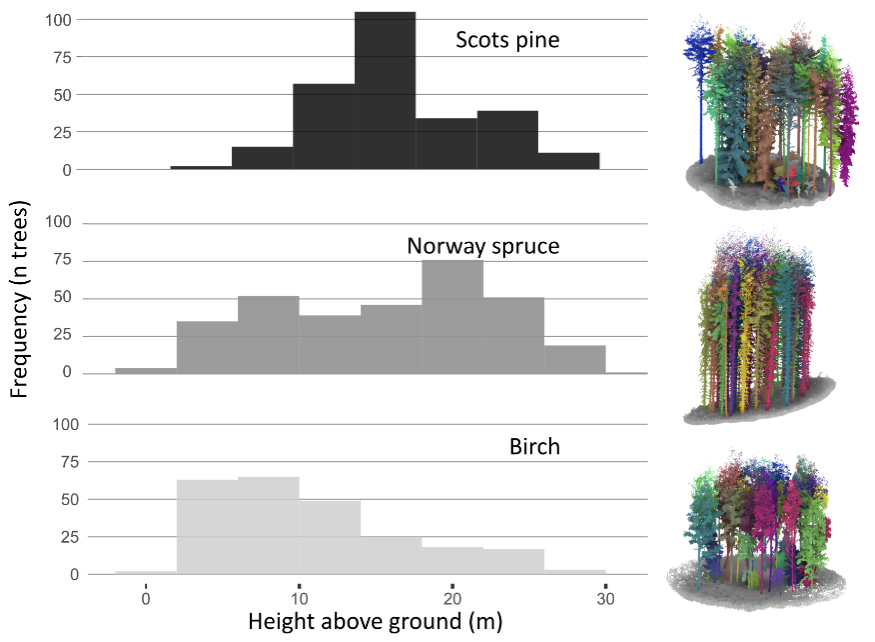}
    \caption{The difference in forest structures included in the sample plots by tree species. \label{fig:forest-structure}}
\end{figure}

The study area was located in an actively managed boreal coniferous dominated forest in Våler municipality in south-eastern Norway (N~\ang{59.503219}, E~\ang{10.884240}). A total of 16 circular plots of \SI{400}{\square\meter} were purposefully selected to cover a range of forest conditions in terms of species composition (see Fig.~\ref{fig:forest-structure}) and stand density (\numrange{200}{2500} trees \SI{}{\per\hectare}; Tab.~\ref{tab:plot-summary}). These plots included forests where the dominant species was either Norway spruce (\textit{Picea abies} (L.) Karst.), Scots pine (\textit{Pinus sylvestris} L.), or birch (\textit{Betula pubescens} or \textit{Betula pendula}), including different degrees of mixing between the species. Concerning the developmental stages, the selected plots were located either in mature forests stands or in stands in the middle of their rotation period. However, no young forest in the regeneration phase were included. 

\begin{table}[htbp]
    \centering
    \begin{tabular}{lcccc}
    \toprule
         & min & mean & max & sd \\
    \midrule
       trees \SI{}{\per\hectare} & 200 & 1298 & 2500 & 638 \\
       dominant height [\SI{}{\meter}] & 15.6 & 22.4 & 29.0 & 4.1 \\
    \bottomrule
    \end{tabular}
    \caption{Summary statistics for the selected plots.\label{tab:plot-summary}}
\end{table}

\subsection{MLS data acquisition}
\Gls{MLS} data was collected in June 2022 using a GeoSLAM ZEB‐HORIZON \citep{geoslam2020manual} in correspondence to the \num{16} circular field plots of \SI{400}{\square\meter} area. The data collection was initialized by booting the GeoSLAM ZEB‐HORIZON in the center of the field plot. Consequently, the operator was walking two perpendicular eight figures extending for a diameter of approximately \SI{30}{\meter}, followed by a walk around the plot's perimeter. The data collection lasted for \SIrange{10}{15}{\minute} per plot.
The raw \gls{MLS} data were then processed within the GeoSLAM Hub software relying on a proprietary SLAM algorithm. The resulting point clouds were down-sampled to only \SI{9}{\percent} of the total points and exported as .las files. This value is the default value in GeoSLAM Hub software, which in previous experiences was found to reduces data redundancy while maintaining the 3D structure information. The whole point clouds were further clipped to include only the area of the plot plus a buffer of \SI{5}{\meter} around the plot area to ensure that all crowns of trees at the edge of the plots could be segmented.

\subsection{Point cloud annotation}
The point clouds corresponding to the 16 selected plots were manually annotated using CloudCompare \citep{cloudcompare} by a team of two annotators, followed by a review step by the annotators' administrator. The annotation consisted of two consecutive steps: 
(1) Instance annotation: segmentation of single trees if they could be identified as trees (i.e., not always possible for small understory trees). The segmentation was done so that branches of intermingled trees were separated as far as practically possible. 
(2) Semantic annotation: the annotators classified every single point into the following classes: ground, vegetation (branches, leaves, and low vegetation), coarse woody debris (i.e., deadwood), and stems. The classes were the same as those defined by \citet{krisanski2021sensor}, with the difference that the stems were separated from the branches and assigned to a general vegetation class. The reason behind this modification of the semantic classes,  was that the stems are distinct features in coniferous forests that can enable a more precise separation of the single instances.  The points from trees with the stem at breast height outside the plot were removed from further analysis.

\begin{figure}
    \centering
    \includegraphics[width=0.8\textwidth]{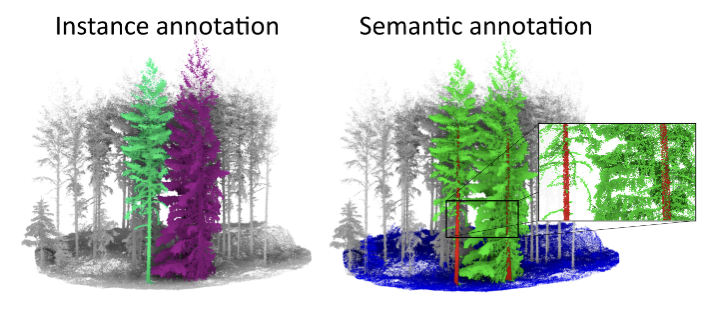}
    \caption{Visualization of the instance and semantic annotation with a detail showing how the stem class was separated from the rest of the crown.\label{fig:instance-vs-semantic}}
\end{figure}

We followed the same split approach by \citet{krisanski2021sensor} and divided the circular plot into four radial slices and randomly assigned two for training (\SI{50}{\percent}), one for validation (\SI{25}{\percent}) and one for the testing (\SI{25}{\percent}). 
A complete overview of the segmented plot can be found in the supplementary materials in Fig.~\ref{fig:plot-visualization}.

\section{Methods}

\begin{figure}[htbp]
    \centering
    \includegraphics[width=1.0\textwidth]{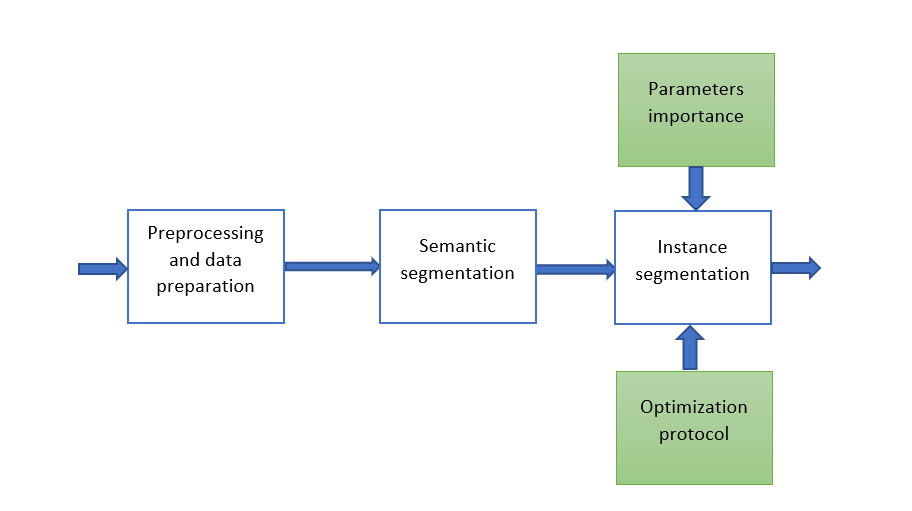}
    \caption{Block diagram of the modular architecture of the processing pipeline.\label{fig:drawing-system}}
\end{figure}

\begin{figure}[htbp]
    \centering
    \includegraphics[width=1.0\textwidth]{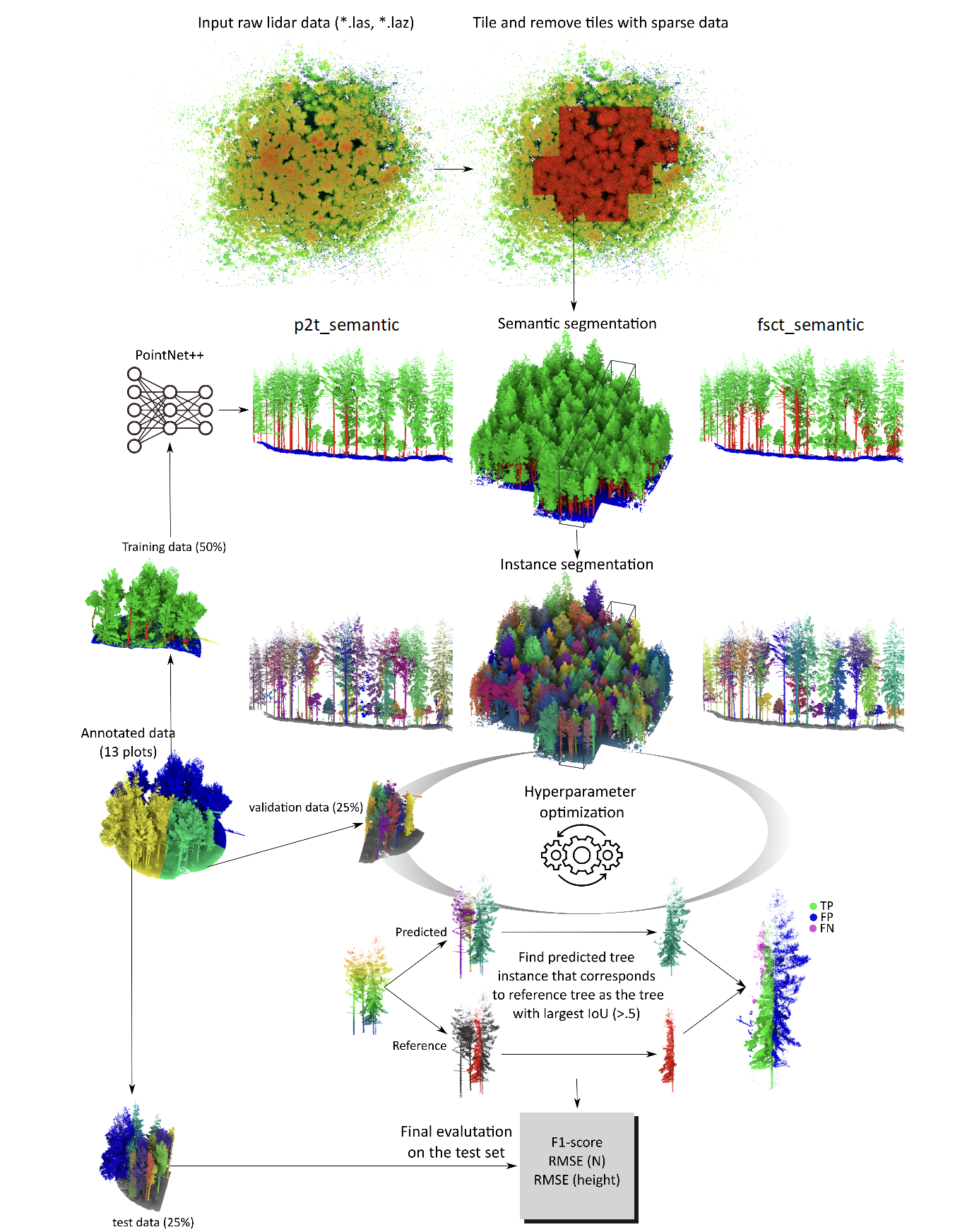}
    \caption{Schematic representation of the adopted workflow.\label{fig:workflow}}
\end{figure}

The pipeline used in this work comprises several stages, as presented in Fig. \ref{fig:drawing-system} and ~\ref{fig:workflow} . It is worth noting that due to the large size of point cloud files, arranging all the steps in the pipeline well and orchestrating their behavior to obtain a good system performance is essential. In some cases, it may be hard to arrive at the processing results due to ineffective data processing which do not account for all the aspects of the data well. This effect may mostly occur when dealing with locally very sparse point clouds (often on borders of the cloud). Therefore, all steps are prepared in a modular fashion and the stages are parameterized  so they can be adjusted to a different densities and types of point clouds.  Furthermore, the steps of the pipeline are prepared in the way which enables smooth substitution of selected components of the system. The elements of the system are interconnected using programming language agnostic composition strategy as presented in Fig. \ref{fig:drawing-system}.

Within this work we used a custom naming convention.  Tab. \ref{tab:pipeline-structure} presents a set of acronyms and features of different pipelines we partially based our framework and which we compare against our performance. Point2Tree framework is equipped with optimization, analytics modules and enables also incorporation of external module implemented in different programming languages thanks to its modular architecture. This feature of P2T was used in order to employ semantic segmentation from FSCT and Instance segmentation from TLS pipelines.

\begin{table}[h]
\centering
\begin{tabular}{|p{3.5cm}|p{2.5cm}|p{2.5cm}|p{2.5cm}|}
\hline
\textbf{Components of the pipeline} & \textbf{FSCT \citep{krisanski2021sensor}} & \textbf{TLS2trees \citep{wilkes2022TLS2trees}} & \textbf{Point2Tree (P2T)} \\
\hline
Semantic seg. & fsct\_semantic & tls\_semantic & p2t\_semantic* \\
Instance seg. & fsct\_instance & tls\_instance & tls\_instance \\
Optimization & no & no & yes \\
Analytics & no & no & yes \\
Language agnostic modularity & no & no & yes \\
\hline
\end{tabular}
\caption{Components of the pipeline and their properties. *This is fsct\_semantic but trained on our data.}
\label{tab:pipeline-structure}
\end{table}

It is worth noting that (see Tab. \ref{tab:pipeline-structure}) that "p2t\_semantic" was obtained from "fsct\_semantic" by replacing original set of FSCT weights of Pointnet++ model with the model trained on our data.

\subsection{Data preprocessing}
In the preprocessing step, the initial tiling operation is done. Tile size is adjustable and should be chosen based on the data profile. It is also worth noting that there is usually an entire range of low-density point cloud tiles on in the edges of point clouds. This data is difficult to digest for both semantic segmentation and instance extraction part of the pipeline. Therefore,  a dedicated procedure to remove these low-density tiles was adopted. 
The procedure examines all the tiles in terms of their point density. A tile is removed from further processing if the density is below a critical threshold. As may be noticed, the tiles' granularity strictly affects the point cloud's shape after the protocol. The effect of the density-based tiling protocol is presented in Fig.~\ref{fig:workflow}.

\subsection{Semantic segmentation}
In our flow Pointnet++ \citep{qi2017pointnetplusplus,krisanski2021sensor} implemented in Pytorch was used as a base model for semantic segmentation. The model was trained from scratch using the newly annotated dataset (p2t\_semantic,  see Tab.\ref{tab:pipeline-structure}). 

The point cloud was sliced into cube-shaped regions to prepare the data for Pointnet++. Each cube was shifted to the origin before inference to avoid floating-point precision issues. The preprocessing is performed before training or inference, and each sample is stored in a file to minimize computational time and facilitate taking advantage of parallel processing. The preprocessing also takes advantage of vectorization by using the NumPy package.

During training, we used subsampling and voxelization protocol to \SI{1}{\centi\meter}. 
The parameters used in the training process are listed in Tab.~\ref{tab:params_sem}, and the training was done on Nvidia GV100GL [Tesla V100S PCIe 32GB]. 

\begin{table}[htbp]
\centering
\begin{tabular}{cc}
    \toprule
    \textbf{parameter name} & \textbf{value} \\
    \midrule
    $\mathrm{num\_epochs}$ & $300$ \\
    $\mathrm{learning\_rate}$ & $0.00005$ \\
    $\mathrm{input\_point\_cloud}$ & $\mathrm{None}$ \\
    $\mathrm{sample\_box\_size\_m}$ & $[6, 6, 8]$ \\
    $\mathrm{sample\_box\_overlap}$ & $[0.5, 0.5, 0.5]$ \\
    $\mathrm{min\_points\_per\_box}$ & $1000$ \\
    $\mathrm{max\_points\_per\_box}$ & $20000$ \\
    $\mathrm{subsample}$ & $\mathrm{True}$ \\
    $\mathrm{subsampling\_min\_spacing}$ & $0.01$ \\
    \bottomrule
\end{tabular}
\caption{The parameters used in the training process for p2t\_sementic module.\label{tab:params_sem}}
\end{table}

\subsection{Instance segmentation}
We employed the \textit{TLS2trees} instance segmentation technique \cite{wilkes2022TLS2trees}.
This segments trees through a series of steps that follow use initial semantic segmentation as input. 
The \textit{TLS2trees} method initially constructs a graph through the wood classified points. 
A comprehensive explanation of this approach is provided in \cite{wilkes2022TLS2trees}. 
It is important to note that the accuracy of the instance segmentation is reliant on the results of the semantic segmentation as well as the quality of the data employed for training this model.

\subsection{Evaluation}
Evaluating machine learning models and pipelines can be challenging, as it requires providing an appropriate set of metrics and a protocol for applying them in a repeatable and reliable way. While it is much easier to develop a protocol if the point matching at the cloud level is ensured, there is still a way to compare results against the ground truth when that is not the case. 

In the solution presented in this paper, the input data is down-sampled during the pre-processing, leaving only a single random point per voxel. As a result, the number of output points is lower than the input points. There are also small distortion introduced related to several point-cloud conversion in a process of mapping between data formats. We provide a methods based on KNN algorithm for point matching. Our method consists of the following steps:

\begin{enumerate}
\item An algorithm for iterative tree elimination:
\begin{enumerate}
\item Find the biggest trees in the GT.
\item Find the biggest overlap in PD.
\item Assign GT (Ground Truth) to PD (predicted) and eliminate PD.
\item Add to collection (dictionary).
\end{enumerate}
\item Compute tree-level metrics based on the dictionary.
\end{enumerate}

We aggregate results on multiple levels and use a set of common metrics such as F1-score and IOU (Jaccard index) to assess the performance of the model on a pixel level.

\begin{equation}
    \text{Precision} = \frac{\text{True Positives (TP)}}{\text{True Positives (TP)} + \text{False Positives (FP)}}
\end{equation}

\begin{equation}
    \text{Recall} = \frac{\text{True Positives (TP)}}{\text{True Positives (TP)} + \text{False Negatives (FN)}}
\end{equation}

\begin{equation}
    \text{F1 Score} = 2 \cdot \frac{\text{Precision} \cdot \text{Recall}}{\text{Precision} + \text{Recall}}
    \label{eq:f1-score}
\end{equation}

\begin{equation}
    \text{IoU (Jaccard index) } = \frac{\text{Area of Overlap}}{\text{Area of Union}}
\end{equation}

Also the residual height operating on a tree level (Eq. \ref{eq:residual-hight}) as the difference between ground truth height (gt) and predicted height (pred) of trees is calculated.
\begin{equation}
    h_{\text{res}} = h_{\text{gt}} - h_{\text{pred}}
    \label{eq:residual-hight}
\end{equation}

The square root of the average squared difference between ground truth heights (h\_gt) and predicted heights (h\_pred) over a dataset is given by Eq. \ref{eq:rmse-hight}
\begin{equation}
    \text{RMSE} = \sqrt{\frac{1}{N} \sum_{i=1}^{N} (h_{\text{gt}_i} - h_{\text{pred}_i})^2}
    \label{eq:rmse-hight}
\end{equation}

For large datasets, serial execution of the metrics is pretty slow; thus, a parallel version was implemented and used for experiments in this work.

\subsection{Optimization}
This work proposes an optimization protocol based on a Bayesian approach \cite{brochu2010tutorial,shahriari2016taking,snoek2012practical}. It is a sequential method that gradually explores the space of hyperparameters, focusing on the most promising manifolds within it. This method is especially suitable for applications where each iteration is time-consuming, as is the case with processing a large volume of point cloud data, presented in this work. In particular, we compute F1-score (Eq. \ref{eq:f1-score}) as a function of point cloud instance segmentation. 

The method works by constructing a probabilistic model, typically a Gaussian Process (GP), to represent the unknown function and then using an acquisition function to balance exploration and exploitation when deciding on the next point to sample. The objective is to find the global optimum with as few evaluations as possible. The GP is defined by a mean function $\mu(\mathbf{x})$ and a covariance function $k(\mathbf{x}, \mathbf{x}')$, which together describe the function's behavior. The choice of kernel function is crucial for the performance of Bayesian optimization as it encodes the prior belief about the function's smoothness. A commonly used kernel is the squared exponential kernel:

\begin{equation}
k(\mathbf{x}, \mathbf{x}') = \sigma^2_f \exp \left(-\frac{(\mathbf{x} - \mathbf{x}')^2}{2l^2} \right)
\end{equation}

where $\sigma^2_f$ represents the signal variance and $l$ is the length scale parameter, $\mathbf{x}, \mathbf{x}'$ are input vectors for which we want to compute the covariance. They are multi-dimensional and model tree hyperparameters setup.

In the Bayesian optimization framework, we start with a prior distribution over the unknown function, and after each evaluation, we update our beliefs using Bayes' rule. This results in a posterior distribution,which is used to guide the search for the global optimum \citep{brochu2010tutorial}. 

\begin{figure}
    \centering
    \includegraphics[width=0.8\textwidth]{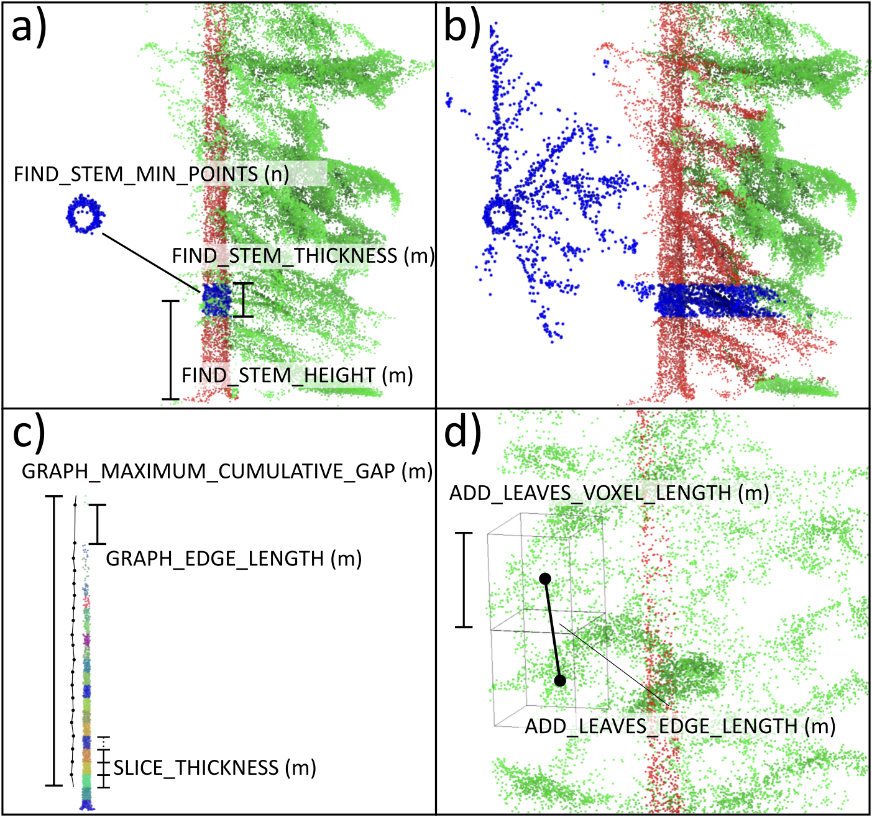}
    \caption{Schematic representations of the meaning of each different tls\_instance hyperparameter tested in this study, including hyperparameters related to the identification of the single tree instances both for p2t\_semantic (a) and fsct\_semantic  (b) semantic segmentation models; the drawing of the stem/wood instance graph (c); and the drawing of the leaves graph (d). \label{fig:optimization}}
\end{figure}

The instance segmentation stage is composed of multiple modules which contain a series of hyperparameters that should be optimized to reach the best possible performance of the model. The most important ones are depicted in Fig.~\ref{fig:optimization}, and tested values are shown in Tab.~\ref{tab:params_optim}.

\begin{table}[htbp]
    \centering
    \begin{tabular}{ll}
        \toprule
        \textbf{parameter} & \textbf{values} \\
        \midrule
        $\mathrm{slice\_thickness}$ & $[0.25, 0.5, 0.75]$ \\
        $\mathrm{find\_stems\_height}$ & $[0.5, 0.75, 1.0, 1.5, 2.0]$ \\
        $\mathrm{find\_stems\_thickness}$ & $[0.25, 0.5, 0.75]$ \\
        $\mathrm{graph\_maximum\_cumulative\_gap}$ & $[1, 2, 3, 4]$ \\
        $\mathrm{add\_leaves\_voxel\_length}$ & $[0.25, 0.5, 0.75]$ \\
        $\mathrm{find\_stems\_min\_points}$ & $[10, 20, 30, 50, 100, 150, 200]$ \\
        \bottomrule
    \end{tabular}
    \caption{Optimization parameters and their ranges. See Figure 5 and Wilkes et al. 2022 for parameter definitions \label{tab:params_optim}}
\end{table}

The chosen values of the hyperparameters cover the most promising and useful ranges. It is worth noting that the choice and the number of parameters affects the performance of the optimization algorithm. Consequently, they should be picked according to the specific profile of the forest dataset in question.  

The optimization process of Point2Tree involves many iterations of the pipeline execution with a distinct set of parameters. Therefore, applying a well-structured protocol to address this process is reasonable.

In each iteration of the optimization process, the F1-score is derived from the complete dataset, and the algorithm maximizes its value over the steps of the execution.

The optimization is done by optimizing the F1-score for the entire set. The overall F1-score is calculated using a three-fold protocol:
\href{}{}
\begin{algorithm}
\caption{F1-score calculation}
\begin{algorithmic}[1]
\For {plot in dataset}
    \For {for tree in  plot}
        \State Compute F1-score
    \EndFor
    \State Aggregate F1-score per plot
\EndFor
\State Aggregate F1-score per dataset
\end{algorithmic}
\end{algorithm}

Based on the F1-score the optimization algorithm guides the next steps of the optimization. In our research and experiments we have noticed that it is possible to improve the optimization results by decomposing the process into several stages. After the initial stage (e.g. 40 runs)  it is possible to stop the optimization and restart it for a limited and the most important set of parameters.

\begin{algorithm}
\caption{Optimization Algorithm - two stage protocol}
\label{alg:opt-two-stages}
\begin{algorithmic}[1]
\Require Initial parameters
\State Run initial optimization
\State Select less then 4 parameters of the highest importance
\State Run optimization for the selected parameters
\Ensure Optimized parameters
\end{algorithmic}
\end{algorithm}

Point2Tree provides a module for assessing the importance of hyparameters in the optimization process. The results are presented in the supplementary material.

\subsection{Final validation}
After completing the optimization, we evaluated the best set of hyperparameters for both Point2Tree (P2T)  with fsct\_semantic  and p2t\_semantic. To validate our results, we compared them against the LAUTx dataset, a benchmark for the instance segmentation \citep{tockner2022lautx}.

Aside from the F1-score, we evaluated additional metrics, including precision, recall, residual height, detection, commission, and omission rate. Finally, we provided comparisons for the regular model with standard parameters used in \cite{wilkes2022TLS2trees} and the optimized set of parameters.

\section{Results and discussion}

\subsection{Semantic segmentation performance}
\begin{figure}
    \centering
    \includegraphics[width=0.7\textwidth]{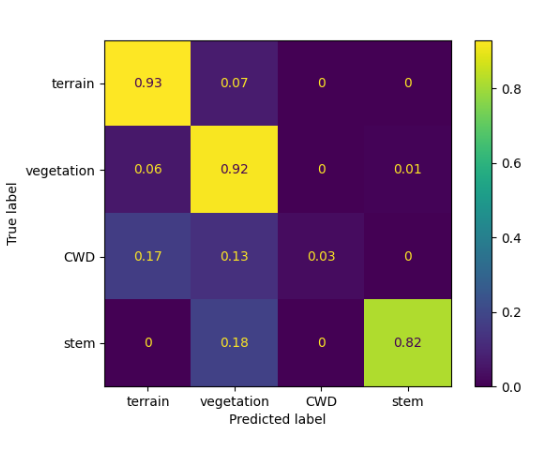}
    \caption{Confusion matrix for p2t\_semantic model}
    \label{fig:confusion-matrix-ours}
\end{figure}

The newly trained p2t\_semantic model achieved precision, recall, and an F1-scores of 0.92. The F1-score ranking per class (vegetation, terrain, and stem) reflected the proportions of these classes in the training data, as shown in Table~\ref{tab:num-points-sem-train}. Vegetation was the most common class, representing over 61\% of the dataset, while CWD and stem were the least common, accounting for only 0.34\% and 13.6\% of the dataset, respectively. The confusion matrix for p2t\_semantic model is shown in Figure~\ref{fig:confusion-matrix-ours}, while Table~\ref{tab:precision-per-class} shows the class-wise metrics for semantic segmentation.

\begin{table}
\centering
\begin{tabular}{ccc}
\toprule
\textbf{Class} & \textbf{Count} & \textbf{Percentage (\%)} \\
\midrule
Terrain & 16,460,251 & 24.94 \\
Vegetation & 40,332,257 & 61.12 \\
CWD & 225,093 & 0.34 \\
Stem & 8,972,140 & 13.60 \\
\bottomrule
\end{tabular}
\caption{Number of point-cloud points in the dataset used for training semantic segmentation model p2t\_semantic}
\label{tab:num-points-sem-train}
\end{table}

\begin{table}[htbp]
    \centering
    \begin{tabular}{lccc}
    \toprule
        & \multicolumn{3}{c}{\textbf{class}} \\
        \cmidrule{2-4}
        \textbf{metric} & \textbf{terrain} & \textbf{vegetation} & \textbf{stem} \\ \midrule
        Precision & 0.86 & 0.94 & 0.94 \\ 
        Recall & 0.93 & 0.92 & 0.82  \\ 
        F1-score & 0.89 & 0.93 & 0.87  \\ 
        \bottomrule
    \end{tabular}
    \caption{Class-wise metrics for p2t\_semantic semantic segmentation}
    \label{tab:precision-per-class}
\end{table}

\subsection[Instance segmentation]{Instance segmentation optimization}
We applied the Bayesian flow for the optimization of hyperparameters for the instance segmentation  \cite{brochu2010tutorial} and considered all the hyperparameters as presented in Tab. \ref{tab:params_optim}. We optimized the hyperparameters for (1) TLS2Trees \cite{wilkes2022TLS2trees} that uses the semantic segmentation from FCST i.e. fsct\_semantic 
  \cite{krisanski2021sensor} and (2) Point2Tree which uses the semantic segmentation model developed in this paper and the instance segmentation framework from TLS2Trees i.e. tls\_instance.  

\begin{figure}[htbp]
    \centering
    \includegraphics[width=0.9\textwidth]{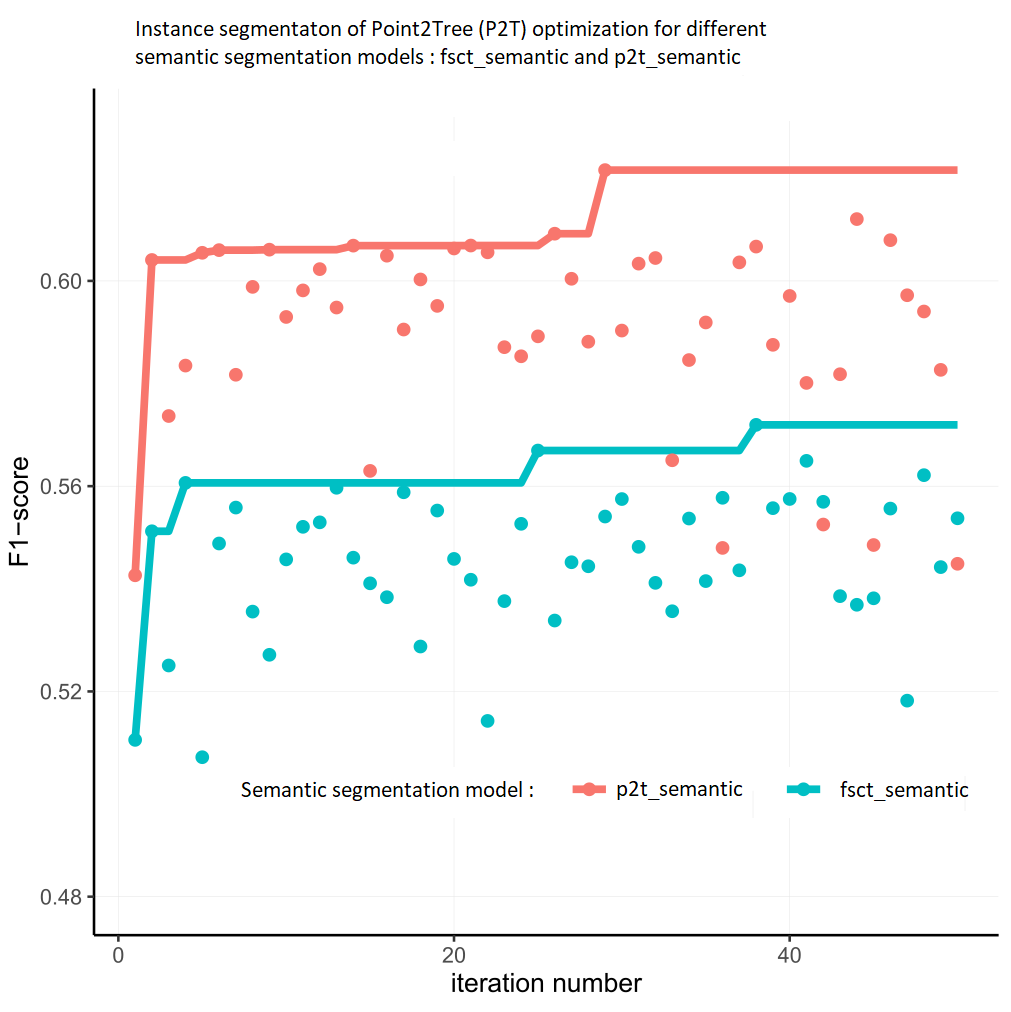}
    \caption{F1-score across the optimization iterations of instance segmentation with two different semantic segmentation models i.e. fsct\_semantic and p2t\_semantic}
    \label{fig:f1-score-iterations}
\end{figure}

The optimization was interrupted after 50 iterations as only minor marginal improvements were observed for both tested models (see Fig.~\ref{fig:f1-score-iterations}). Interestingly, when considering all iterations, the rate of growth of the F1-score over the number of iterations was negative for the optimization of Point2Tree (slope of -0.0002). At the same time, it was slightly positive TLS2Trees (slope of 0.0002). These numbers indicate that the method Point2Tree was more robust to variations in the choice of hyperparameters. Such property is desired and might be explained by the definition of the instances being more robust when based only on clean stems rather than using a class merging stems and woody branches. 

The analysis of the importance of the different hyperparameters on the F1-score revealed differences between the two approaches (see Fig. \ref{fig:importance_comparison}). In particular, we found the following for the respective hyperparameters: 
\begin{itemize}
    \item $\mathrm{find\_stem\_height}$: there was a contrasting effect between the Point2Tree with p2t\_semantic and fsct\_semantic. In the latter, $\mathrm{find\_stem\_height}$ was the important hyperparameter, while it was the least important for p2t\_semantic  approach. Point2Tree results showed that fsct\_semantic  semantic segmentation approach selected higher (approx. 1.75 m above ground) slices, whereas p2t\_semantic  preferred close-to-the-ground (approx. 0.8 m) slices. 
    The need for  fsct\_semantic to search for tree instances higher up the stem might be due to larger noise due to low branches and low vegetation in the wood instance class used for clustering the instances. In this context, p2t\_semantic proved more robust in filtering out low-vegetation and non-stem points.

    \item $\mathrm{find\_stems\_thickness}$: for this hyperparameter, defining the thickness of the slice used for clustering the instances, the two approaches also behaved differently, with P2T-fsct\_semantic approach tended to select narrower (approx. 20 cm) slices, whereas P2T with p2t\_semantic  approach preferred wider (approx. 0.75 m) slices. The selection of narrower search windows in P2T-fsct\_semantic approach, coupled with the selection of higher slices, is needed to reduce the noise due to branches and low vegetation. Using wider slices allows for including more extensive portions of a tree stem, thus increasing the chance of detecting single instances.

    \item $\mathrm{find\_stem\_min\_points}$: this was the most crucial hyperparameter for P2T-p2t\_semantic approach and was negatively correlated with the F1-score (-0.4), meaning that the minimum number of points to trigger a new instance was 50-100 points, rather than 200 points in the P2T-fsct\_semantic default values. On the other hand, for the P2T-fsct\_semantic approach, this hyperparameter was the second least important and negatively correlated with the f1-score(-0.26), preferring values between 100 and 120 points.  

    \item $\mathrm{graph\_edge\_length}$: This was the second most important hyperparameter for P2T-p2t\_semantic approach but was very weakly correlated (0.008) with the F1-score. The preferred value for P2T-pt2\_semantic model was around 1 m, the same as the default value in TLS2trees. For P2T-fsct\_semantic approach, the most suitable values were in the 0.4 - 0.6 m range. This is the maximum length an edge in the graph can be. If this is set to a larger value then disconnected points (occlusion) can be connected although this may bridge gaps between trees. It relates to the flexibility of the graph growth, and P2T-fsct\_semantic approach is more rigid.

    \item $\mathrm{graph\_maximum\_cumulative\_gap}$ We can see that (Fig. \ref{fig:importance_comparison}) this hyperparameter has a contrast effect on P2T-p2t\_semantic and P2T-fsct\_semantic approaches. In P2T-pt2\_semantic approach, it has a pretty high positive correlation (0.25) and kind of medium importance, which means that long gaps are accepted, resulting in more additional stems and branches to be included (see \cite{wilkes2022TLS2trees} Fig. 7). This is acceptable and desirable in P2T-p2t\_semantic case since we skip branches and focus only on trunks. In P2T-fsct\_semantic method, the smaller values of $\mathrm{graph\_maximum\_cumulative\_gap}$ are preferred, which may be considered an attempt to reduce the noise in the form of small branches and stems. The parameter is also important in P2T-fsct\_semantic.

    \item $\mathrm{add\_leaves\_voxel\_length}$  In the case of $\mathrm{add\_leaves\_voxel\_length}$ again the impact contrasting. In the case of P2T-p2t\_semantic approach, the parameter is quite important (approx. 016) and slightly positively correlated. Those values indicate that the voxel size affects the output F1-score of our method. This relation is expected because our method is based on trunk modeling without branches, so the size of the leaf voxel is important. On the other hand, attempting to manipulate this parameter does not lead to high gains in model performance. Conversely, this parameter has a negative (approx. -0.35) correlation in the P2T-fsct\_semantic method but even smaller importance. This lack of impact can be explained by the fact that since the P2T-fsct\_semantic approach is based on the branches, the size of the leaf voxels is not that critical, and also, it is preferred to be low to give more freedom to the graph constructing algorithm in TLS. 
    
    \item $\mathrm{add\_leaves\_edge\_length}$ This parameter, in both the approaches, has a low correlation and low importance. The overall impact of the parameter on the output is relatively low.

    \item $\mathrm{slice\_thickness}$ This parameter is unimportant in both methods. However, it is worth noting that in the case of P2T-fsct\_semantic method, it has a more negative correlation. Therefore, in P2T-fsct\_semantic approach, it is better to have smaller slices, which can be beneficial since the algorithm operates on branches. On the other hand, in P2T-p2t\_semantic approach, trunks are more distinguishable as an effect of the lack of branches, so the TLS algorithm can allow larger slices, which is reflected in the correlation of the parameter. 

\end{itemize}

More detailed examination of Fig. \ref{fig:importance_comparison} reveals two P2T-fsct\_semantic and five of P2T-p2t\_semantic hyperparameters in the right part of the plot. It means that for P2T-fsct\_semantic approach, there are two important and contrasting parameters, namely $\mathrm{graph\_maximum\_cumulative\_gap}$ and  $\mathrm{find\_stem\_height}$, whereas, for P2T-p2t\_semantic method, there are five of them. Furthermore, in the case of P2T-p2t\_semantic approach, the optimization landscape is blurry since we need to manipulate five less contrasting parameters. 

\subsection{Metrics evaluation}
\subsubsection{On test data from this study}
The evaluation of the metrics computed against the test data (see Tab.~\ref{tab:overall}) revealed that using optimal parameters resulted in a performance boost compared to using default values for both P2T with p2t\_semantic and fsct\_semantic 
. In line with the optimization findings, the magnitude of the improved performance for  P2T-fsct\_semantic was twice as large (0.08 F1-score increase) compared to  P2T-p2t\_semantic (0.04 F1-score increase). When optimized, both approaches reached similar levels of F1-score. Despite the marginal differences,  P2T with p2t\_semantic approach resulted in a smaller residual height of 0.87 m and 3.47 RMSE and a lower detection rate than  P2T-fsct\_semantic. The analysis of the false positive rates indicated that both approaches tended to over-segment the point clouds, and such behavior was more prominent in  P2T-p2t\_semantic. On the other hand, it is essential to highlight that while having more significant commission errors,  P2T-p2t\_semantic model reduced to nearly zero the false negatives (i.e., omitted trees).

\begin{table}[htbp]
    \centering
\begin{tabularx}{\textwidth}{>{\raggedright\arraybackslash}X>{\centering\arraybackslash}m{2cm}>{\centering\arraybackslash}m{1.8cm}>{\centering\arraybackslash}m{2cm}>{\centering\arraybackslash}m{1.8cm}}
\toprule
\textbf{evaluation metric} & \textbf{ P2T-p2t\_semantic} & \textbf{optimized  P2T-p2t\_semantic} & \textbf{ P2T-fsct\_semantic} & \textbf{optimized  P2T-fsct\_semantic} \\
\midrule
Precision & 0.6 & 0.65 & 0.57 & 0.678 \\ 
Recall & 0.61 & 0.65 & 0.57 & 0.644 \\ 
F1-score & 0.57 & 0.61 & 0.54 & 0.625 \\ 
Intersection over Union (IoU) & 0.47 & 0.5 & 0.45 & 0.514 \\ 
Residual Height (gt - pred) [m] &  & 0.87 &  & 0.907 \\ 
Root Mean Square Error (RMSE) of Height Errors [m] &  & 3.47 &  & 3.597 \\ 
True Positive (detection rate) &  & 0.57 &  & 0.594 \\ 
False Positive (commission) &  & 0.36 &  & 0.262 \\ 
False Negative (omissions) &  & 0.07 &  & 0.144 \\ \bottomrule
\end{tabularx}
\caption{Instance segmentation results on the test dataset.}
   \label{tab:overall}
\end{table}

\begin{table}[htbp]
\centering
\begin{tabular}{lcc}
\toprule
\textbf{parameter} & \textbf{ P2T-fsct\_semantic} & \textbf{ P2T-p2t\_semantic} \\ \midrule
$\mathrm{add\_leaves\_edge\_length}$ & 1.115 & 0.991 \\ 
$\mathrm{add\_leaves\_voxel\_length}$ & 0.159 & 0.317 \\ 
$\mathrm{find\_stems\_height}$ & 1.614 & 0.797 \\ 
$\mathrm{find\_stems\_min\_points}$ & 84 & 105 \\ 
$\mathrm{find\_stems\_thickness}$ & 0.46 & 0.75 \\ 
$\mathrm{graph\_edge\_length}$ & 0.66 & 1.086 \\ 
$\mathrm{graph\_maximum\_cumulative\_gap}$ & 6.826 & 14.845 \\ 
$\mathrm{slice\_thickness}$ & 0.678 & 0.833 \\ \bottomrule
\end{tabular}
\caption{Set of the best parameters for both of the models.}
\label{tab:set-best-param}
\end{table}

The experiments presented in Tab. \ref{tab:overall} were conducted for a set of the best hyperparameters which we obtained in the optimization process. The set is given in Tab.\ref{tab:set-best-param}.

\subsubsection{On the LAUTx data}
The results from applying the  P2T with fsct\_semantic and  P2T with p2t\_semantic pipelines to the LAUTx data for each of the different sets of hyperparameters revealed that both approaches performed consistently or even better than what was found for our initial test data (see Tab.~\ref{tab:lautx}). 

Interestingly, the default parameters were more suitable than the optimized hyperparameters, highlighting the need for more extensive and varied datasets of annotated plots for more robust optimization.  Alternatively, the optimization process can be done separately for each dataset but this requires that a part of a dataset is labeled for the  Point2Tree pipepline for adjusting the hyperparamters of the instance segmentation.

\begin{table}[htbp]
\centering
\begin{tabularx}{\textwidth}{X>{\centering\arraybackslash}m{1.8cm}>{\centering\arraybackslash}m{1.8cm}>{\centering\arraybackslash}m{1.8cm}>{\centering\arraybackslash}m{1.8cm}}
\toprule
& \multicolumn{2}{c}{\textbf{ P2T with p2t\_semantic}} & \multicolumn{2}{c}{\textbf{ P2T with fsct\_semantic}} \\
\cmidrule(r){2-3}\cmidrule(l){4-5}
\textbf{metric} & \textbf{default} & \textbf{optimized} & \textbf{default} & \textbf{optimized} \\ \midrule
Precision & 0.826 & 0.787 & 0.773 & 0.771 \\ 
Recall & 0.588 & 0.528 & 0.57 & 0.506 \\ 
F1-score & 0.669 & 0.603 & 0.629 & 0.578 \\ 
Intersection over Union (IoU) & 0.533 & 0.461 & 0.502 & 0.442 \\ 
Residual Height (gt - pred) [m] & 0.484 & 0.563 & 0.967 & 1.094 \\ 
Root Mean Square Error (RMSE) of Height Errors [m] & 2.66 & 3.496 & 3.589 & 4.971 \\ 
True Positive (detection rate) & 0.552 & 0.422 & 0.405 & 0.366 \\ 
False Positive (commission) & 0.294 & 0.456 & 0.29 & 0.427 \\ 
False Negative (omissions) & 0.154 & 0.122 & 0.304 & 0.207 \\ 
\bottomrule
\end{tabularx}
\caption{Instance segmentation benchmark LAUTx dataset \cite{tockner2022lautx}}
\label{tab:lautx}
\end{table}

The  P2T with p2t\_semantic approach performs slightly better than the  P2T fsct\_semantic. This result may stem from p2t\_semantic  model focusing more on the trunks, and the LAUTx dataset \cite{tockner2022lautx} is composed of high and mostly separated trees.

\begin{figure}
\centering
\includegraphics[width=0.7\textwidth]{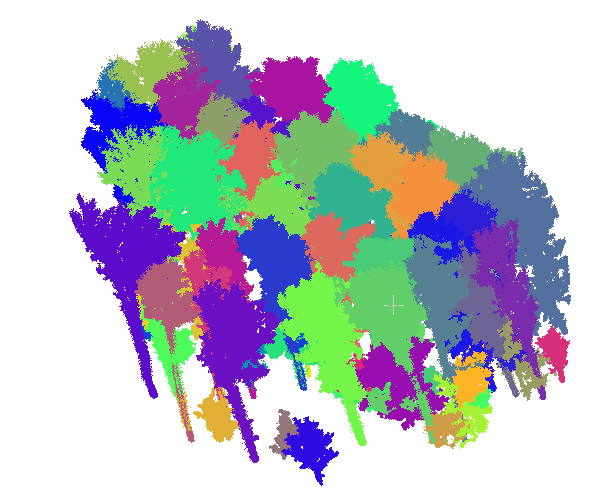}
\caption{Sample instance segmentation with  P2T p2t\_semantic on the benchmark LAUTx dataset \cite{tockner2022lautx} }
\label{fig:sample_austrain_instance_seg}
\end{figure}

Sample results of the performance of the  P2T-p2t\_semantic on the LAUTx dataset \cite{tockner2022lautx} are provided in Fig.\ref{fig:sample_austrain_instance_seg}.

\begin{figure}
\centering
\includegraphics[width=0.7\textwidth]{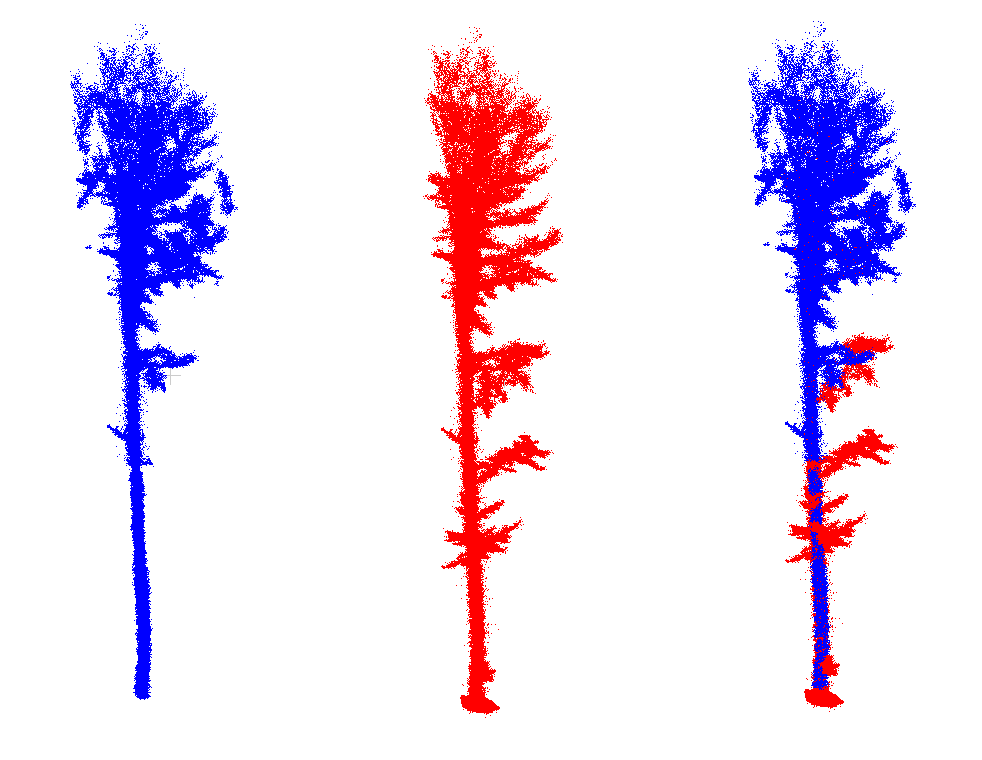}
\caption{Sample gt (red) and predicted (blue) for precision=0.89,	recall=0.69,	f1=0.78	IoU=0.64
\cite{tockner2022lautx} }
\label{fig:sample_gt_pred}
\end{figure}

\begin{figure}
\centering
\includegraphics[width=0.7\textwidth]{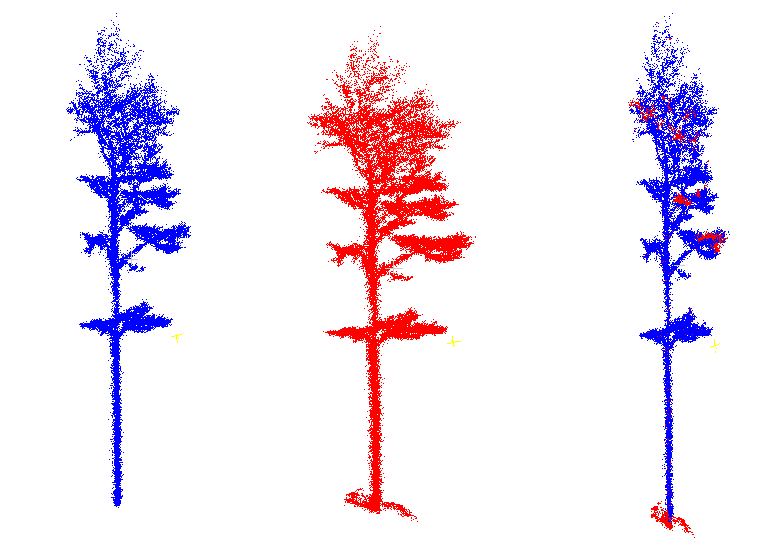}
\caption{Sample gt (red) and predicted (blue) for precision=0.99,	recall=0.90, f1=0.94, IoU=0.89
\cite{tockner2022lautx} }
\label{fig:sample_gt_pred_high}
\end{figure}

\begin{figure}
\centering
\includegraphics[width=0.7\textwidth]{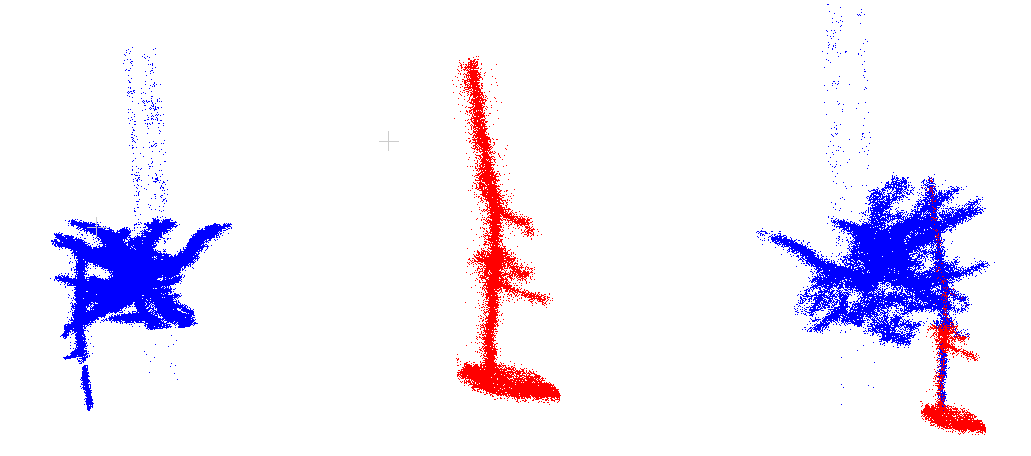}
\caption{Sample gt (red) and predicted (blue) for precision=0.07,	recall=0.33, f1=0.12, IoU=0.06
\cite{tockner2022lautx} }
\label{fig:sample_gt_pred_low}
\end{figure}

\begin{figure}
\centering
\includegraphics[width=\textwidth]{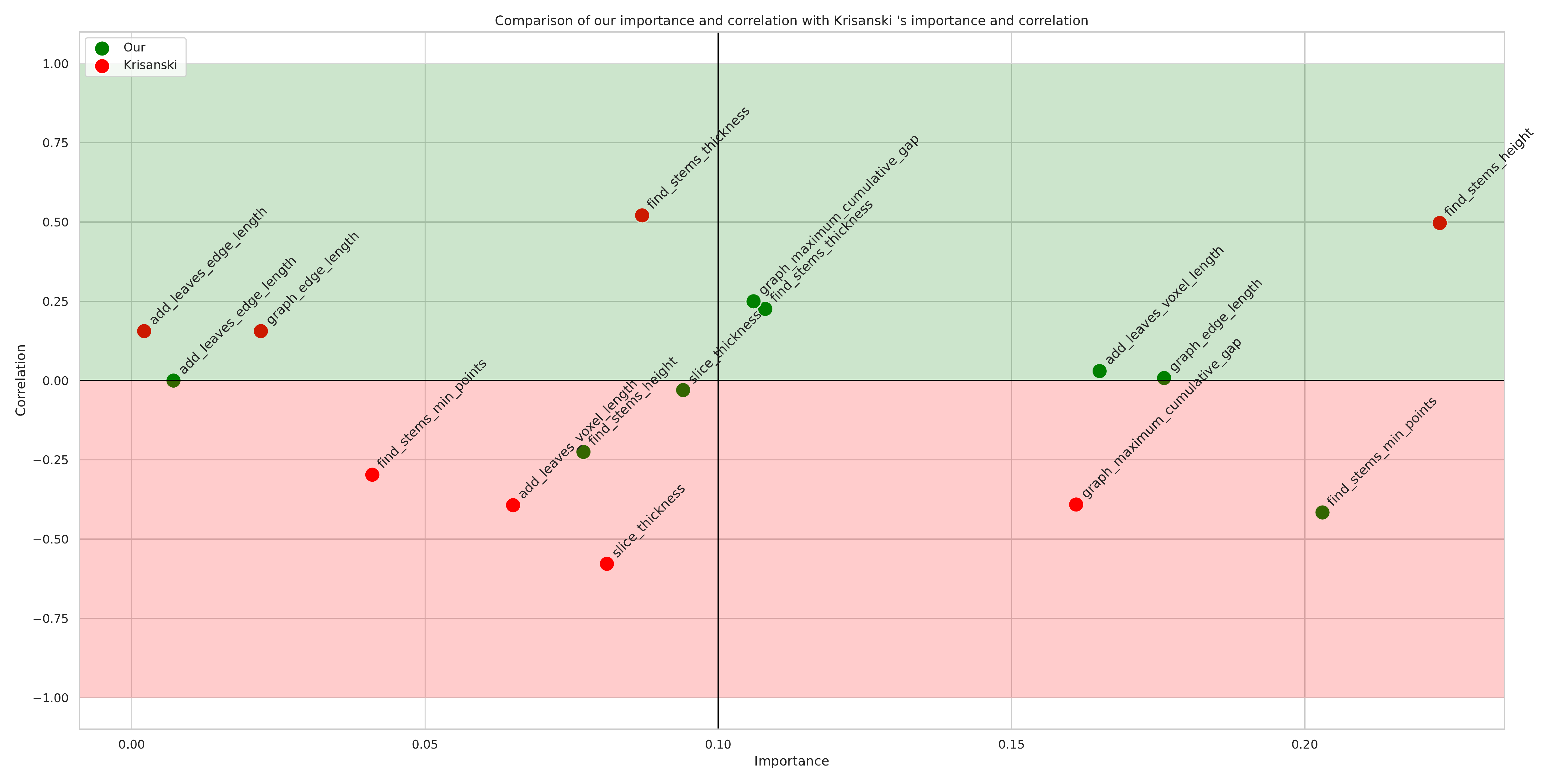}
\caption{Comparison of  P2T with p2t\_semantic parameters importance and correlation with P2T with fsct\_semantic importance and correlation }
\label{fig:importance_comparison}
\end{figure}

In order to visualize the performance of the metrics used in the experiments, we have provided a series of plots for different quality of results. They are given by Fig. \ref{fig:sample_gt_pred}, \ref{fig:sample_gt_pred_low} and \ref{fig:sample_gt_pred_high}. We can see that the number of artifacts grows once the metrics values go down. It is also worth noting that some artifacts are specific to a given dataset labeling process. For instance, as we can see in Fig. \ref{fig:sample_gt_pred}, \ref{fig:sample_gt_pred_low} and \ref{fig:sample_gt_pred_high}, the bottom of tree trunks were very deeply labeled, such that the labeling account for a part of the ground. This issue is not the case for the data p2t\_semantic model was trained on. Thus it is hard to achieve a perfect metric on the LAUTx dataset \cite{tockner2022lautx}. 

    \section{Conclusions}

The study show a new framework for point cloud instance segmentation. The framework consist of series of components which are structured in the flexible way. This allows to replace a selected parts of the pipeline (e.g. instance segmentation) with new or alternative modules. The new modules may be implemented in alternative languages (e.g. C++, java etc.) and still the integration is possible. The framework is also equipped with optimization module and important parameters visulization.

We also tested the effect of hyperparameter tuning the TLS2trees instance segmentation pipeline, developed initially mainly for tropical forests, to optimal settings for coniferous forests. Further, we tested the effect of using a semantic segmentation model specifically designed to focus on identifying the stems of coniferous trees on the tree instance segmentation accuracy. Our study found that the hyperparameter tuning positively affected the segmentation output quality of our data. However, when applying the same parameters to the external LAUTx dataset, the performance was poorer than using the default setting. This result indicates that to estimate a more robust and transferable set of hyperparameters, we need to develop more extensive databases of openly available annotated point cloud data spanning a broader range of forest types than those used in this study. When optimized, the effect of using different semantic segmentation models (i.e.,  P2T with p2t\_semantic and fsct\_semantic) was marginal. However, it is also true that the instance segmentation relying on p2t\_semantic model seemed less sensitive to the choice of hyperparameters and thus more robust in dense forests or forests with many low branches (e.g., non-self-pruning species).

Due to the architecture of the instance segmentation algorithm there are set of hyperparamters which are not acceptable and lead to collapse of the pipeline (one or several plots break). This imposed another constrain a choice of the protocol for hyperparameter tuning i.e. Bayesian approach for which the space of the hyperparameters does not have to be convex.

\appendix

\section{Supplementary materials}

\begin{figure}
    \centering
    \includegraphics[angle=90,height=0.9\textheight]{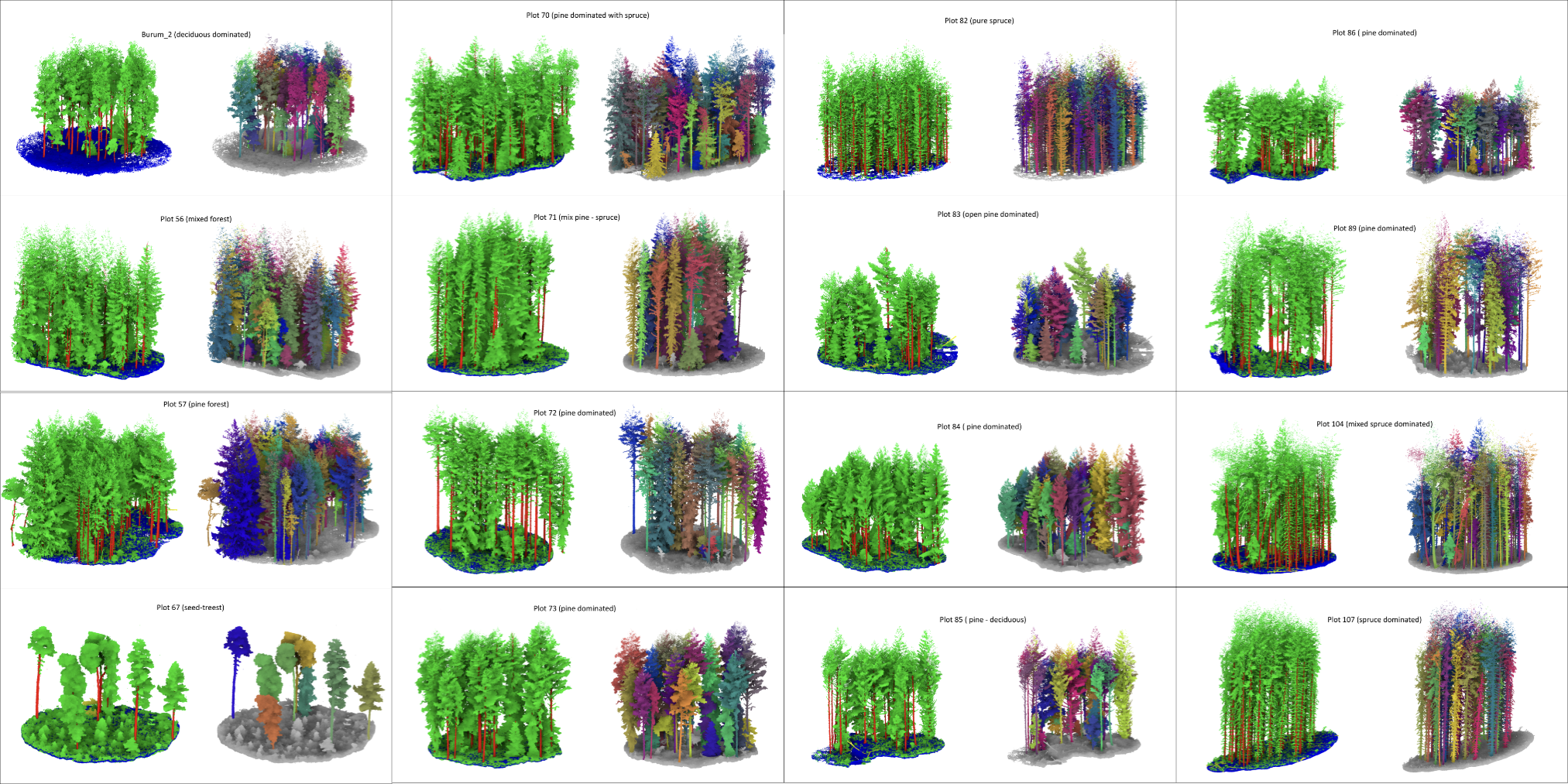}
    \caption{Visualization of the annotated plots for the semantic and instance segmentation.\label{fig:plot-visualization}}
\end{figure}

\begin{table}[htbp]
\begin{tabular}{lcccc}
\toprule
& \multicolumn{2}{c}{\textbf{P2T with p2t\_semantic}} & \multicolumn{2}{c}{\textbf{P2T with fsct\_semantic}} \\
\cmidrule(r){2-3}\cmidrule(l){4-5} 
\textbf{metric} & \textbf{importance} & \textbf{correlation} & \textbf{importance} & \textbf{correlation} \\ \midrule
$\mathrm{find\_stems\_min\_points}$ & 0.2030 & -0.4160 & 0.0410 & -0.2970 \\ 
$\mathrm{add\_leaves\_voxel\_length}$ & 0.1650 & 0.0300 & 0.0650 & -0.3930 \\ 
$\mathrm{graph\_edge\_length}$ & 0.1760 & 0.0080 & 0.0220 & 0.1560 \\ 
$\mathrm{find\_stems\_height}$ & 0.0770 & -0.2250 & 0.2230 & 0.4970 \\ 
$\mathrm{graph\_maximum\_cumulative\_gap}$ & 0.1060 & 0.2500 & 0.1610 & -0.3910 \\ 
$\mathrm{find\_stems\_thickness}$ & 0.1080 & 0.2260 & 0.0870 & 0.5210 \\ 
$\mathrm{slice\_thickness}$ & 0.0940 & -0.0300 & 0.0810 & -0.5780 \\ 
$\mathrm{add\_leaves\_edge\_length}$ & 0.0071 & 0.0000 & 0.0021 & 0.1560 \\ \bottomrule
\end{tabular}
\caption{Importance and correlation of the parameters}
\label{tab:param-importance}
\end{table}

\bibliography{bibliography}

\end{document}